\documentclass[letterpaper, 10 pt, conference]{ieeeconf}  

\IEEEoverridecommandlockouts                              

\usepackage{multicol}
\usepackage{multirow}
\usepackage[bookmarks=true]{hyperref}
\usepackage{booktabs}  
\usepackage{amsmath} 
\usepackage{amssymb}  
\usepackage{graphicx}
\usepackage{pifont}
\usepackage{xcolor}
\usepackage{cuted} 
\usepackage{capt-of}
\usepackage{caption}
\usepackage{eso-pic}

\let\labelindent\relax
\usepackage[inline]{enumitem}

\newcommand{\greentick}{\textcolor{green}{\checkmark}}
\newcommand{\redx}{\textcolor{red}{\text{\sffamily \ding{55}}}}
\graphicspath{ {./figures/} }

\title{\LARGE \bf
$\lambda$: A Benchmark for Data-Efficiency in Long-Horizon Indoor Mobile Manipulation Robotics
\vspace{-4mm}
}

\author{Ahmed Jaafar\textsuperscript{\textdagger$\mathparagraph$},
Shreyas Sundara Raman\textsuperscript{\textdagger},
Sudarshan Harithas\textsuperscript{\textdagger*},
Yichen Wei\textsuperscript{\textdagger*},
Sofia Juliani\textsuperscript{\S}, \\
Anneke Wernerfelt\textsuperscript{\textdaggerdbl},
Benedict Quartey\textsuperscript{\textdagger}, 
Ifrah Idrees\textsuperscript{\textdagger},
Jason Xinyu Liu\textsuperscript{\textdagger}, 
Stefanie Tellex\textsuperscript{\textdagger}
\thanks{\textsuperscript{\textdagger}Brown University, \textsuperscript{\S}Rutgers University, \textsuperscript{\textdaggerdbl}University of Pennsylvania}%
\thanks{\textsuperscript{*}Equal Contribution}%
\thanks{\textsuperscript{$\mathparagraph$}Corresponding author: {\tt\small ahmed\_jaafar@brown.edu}}%
}

\begin{document}


\bstctlcite{BSTcontrol}
\makeatletter
\let\NAT@parse\undefined
\makeatother

\thispagestyle{empty}
\pagestyle{empty}



\makeatletter
\let\@oldmaketitle\@maketitle
\renewcommand{\@maketitle}{\@oldmaketitle
\centering
\includegraphics[width=0.83\textwidth]{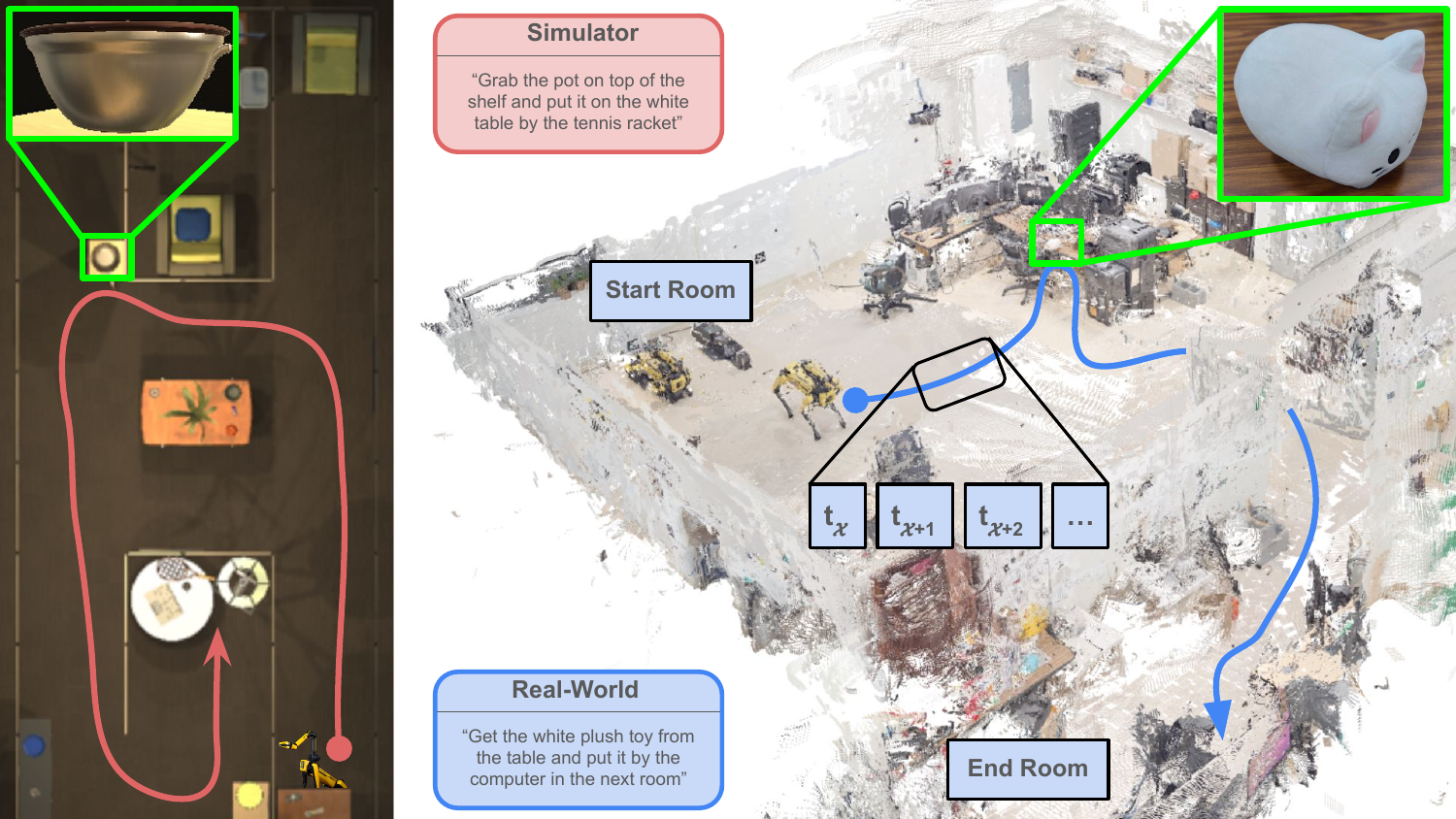}
\captionof{figure}{{\textbf{Overview.} 
    LAMBDA ($\lambda$) is a benchmark of 571 language-conditioned long-horizon mobile manipulation tasks paired with human-collected expert demonstrations in simulated and real-world environments. 
    This figure illustrates an example multi-room task and associated robot trajectory in simulation (left) and real-world (right) environments.
   }
}
\label{fig:main_splash}
\vspace{-5.5mm}
}

\AddToShipoutPictureBG*{%
  \AtTextUpperLeft{
    \raisebox{5mm}{
      \hspace*{0.05\textwidth}
      \parbox{0.9\textwidth}{\centering\small
        © 2025 IEEE.  Personal use of this material is permitted.  Permission from IEEE must be obtained for all other uses, in any current or future media, including reprinting/republishing this material for advertising or promotional purposes, creating new collective works, for resale or redistribution to servers or lists, or reuse of any copyrighted component of this work in other works.
      }%
    }%
  }%
}

\maketitle

\begin{abstract}
Learning to execute long-horizon mobile manipulation tasks is crucial for advancing robotics in household and workplace settings.
However, current approaches are typically data-inefficient, underscoring the need for improved models that require realistically sized benchmarks to evaluate their efficiency. 
To address this, we introduce the \emph{LAMBDA} ($\lambda$) benchmark\footnote{\href{https://lambdabenchmark.github.io/}{\tt\small lambdabenchmark.github.io}}––\underline{\emph{L}}ong-horizon \underline{\emph{A}}ctions for \underline{\emph{M}}obile-manipulation \underline{\emph{B}}enchmarking of \underline{\emph{D}}irected \underline{\emph{A}}ctivities––which evaluates the data efficiency of models on language-conditioned, long-horizon, multi-room, multi-floor, pick-and-place tasks using a dataset of manageable size, more feasible for collection.
Our benchmark includes $571$ human-collected demonstrations that provide realism and diversity in simulated and real-world settings. 
Unlike planner-generated data, these trajectories offer natural variability and replay-verifiability, ensuring robust learning and evaluation. 
We leverage ${\lambda}$ to benchmark current end-to-end learning methods and a modular neuro-symbolic approach that combines foundation models with task and motion planning. 
We find that learning methods, even when pretrained, yield lower success rates, while a neuro-symbolic method performs significantly better and requires less data.



\end{abstract}


\section{INTRODUCTION}
Improving data-efficient learning for long-horizon mobile manipulation (MoMa) tasks is vital for pushing the boundaries of robotics in household and workplace applications.
These tasks, such as fetching an item from a specific location or organizing objects across multiple rooms, often involve understanding language commands, navigating complex spaces, and performing precise pick-and-place actions. 
Strides have been made in tabletop manipulation with various approaches for data-efficient learning~\cite{gem, diffusionpolicy, transporter}; however, existing MoMa models typically rely on large-scale datasets with extensive amounts of trajectories~\cite{rt1, rtx, spoc}. 
Collecting such large-scale datasets is resource-intensive, requiring substantial time, labor, and infrastructure.  
This data inefficiency arises from the inherent challenges of MoMa tasks: extensive observation, action, and state spaces, as well as diverse tasks, and the need for generalization across environments.


Existing benchmarks do not assess models for data efficiency, as they rely on unrealistically large sizes, leaving a significant gap in evaluating progress in this area. 
A few works ~\cite{safeagentbench, infiniteworld, robocasa, m3bench, rt1, partnr} pair task demonstrations with language instruction annotations and support long-horizon tasks~\cite{safeagentbench,behavior1k,infiniteworld,partnr}, providing an intuitive and accessible interface for specifying robot objectives. 
Even fewer~\cite{robocasa, partnr} provide natural language instructions in an open-ended, flexible format rather than relying on fixed templates.
Many benchmarks lack real-world data~\cite{mobilealoha, behavior1k, robocasa, BRMData, demobot, rt1} or rely on planner-generated trajectories~\cite{maniskillhab, m3bench, m2diffuser}, limiting their practicality. 
Quadrupeds excel in diverse terrains~\cite{spot} and are well-suited for long-horizon MoMa tasks, yet only RoboSuite~\cite{robosuite}, RoboCasa~\cite{robocasa}, and DeMoBot~\cite{demobot} provide real quadruped data.

To address these challenges, we propose the \emph{LAMBDA} ($\lambda$) benchmark (\underline{\emph{L}}ong-horizon \underline{\emph{A}}ctions for \underline{\emph{M}}obile-manipulation \underline{\emph{B}}enchmarking of \underline{\emph{D}}irected \underline{\emph{A}}ctivities).
$\lambda$ evaluates models on a realistic amount of demonstrations of language-conditioned, multi-room, multi-floor, pick-and-place tasks requiring long-horizon planning and execution, addressing the limitations of existing benchmarks.
Our dataset consists of $571$ trajectories, which were crowdsourced and manually collected to ensure realism, diversity, and replay-verifiability. 
Crowdsourcing was specifically conducted to capture natural, flowing instructions rather than templated commands, providing a realistic distribution of tasks that people want robots to perform in household and workplace environments. 
By including language-conditioned tasks, $\lambda$ supports intuitive and accessible interactions, making it practical for real-world users.
Unlike planner-generated data, human-collected trajectories in $\lambda$ provide natural variability and expertise, better capturing real-world robot operations.
Additionally, $\lambda$ supports long-horizon tasks involving navigation across rooms and floors, reflecting the realistic demands of indoor scenarios. 
Simulated environments enable diverse testing scenarios and object variability, while real-world data ensures models are prepared for practical deployment. We collected the real data with a quadruped robot due to its optimal form factor, suitable for navigating complex terrain.

We used $\lambda$ to benchmark several models and establish baseline performance for long-horizon MoMa tasks. 
Behavior cloning (BC) models trained on the dataset showed low success rates, such as $2.7\%$ and $5.1\%$, indicating data inefficiency while confirming the dataset’s ability to support learning.
A neuro-symbolic modular approach, combining foundation models with task and motion planning, achieved a $44.4\%$ success rate, significantly outperforming the BC models, illustrating the potential of such approaches to enable more data-efficient MoMa systems.

\textbf{Overall, our contributions are summarized as follows:}
\begin{enumerate*}[label=\textbf{(\arabic*)}]
    \item \textbf{$\lambda$ benchmark:} We introduce a novel benchmark specifically tailored in size to evaluate data efficiency in MoMa, emphasizing language-conditioned, long-horizon pick-and-place indoor tasks in realistic household and workplace settings.
    \item \textbf{High-quality dataset:} $\lambda$'s tasks are paired with $571$ human-collected robot demonstrations, providing natural variability, replay-verifiability, and a balance of simulated and real-world data for robust model evaluation.
    \item \textbf{Comprehensive evaluation:} We benchmark multiple methods, including BC and modular neuro-symbolic approaches, providing a baseline for future MoMa system development.
\end{enumerate*}


\begin{table*}[ht!]
\centering
\renewcommand{\arraystretch}{1}
\resizebox{\textwidth}{!}{%
\begin{tabular}{lcccccccc}
\hline
\textbf{Benchmark}         & \textbf{NL} & \textbf{FF NL} & \textbf{Long-Horizon} & \textbf{Demos} & \textbf{Human-Collected} & \textbf{Sim} & \textbf{Real} & \textbf{Quadruped} \\
\hline
SafeAgentBench~\cite{safeagentbench} & \greentick & \redx & \greentick & \redx & \redx & \greentick & \redx & \redx \\
HumanoidBench~\cite{humanoidbench} & \redx & \redx & \redx & \redx & \redx & \greentick & \redx & \redx \\
Mobile ALOHA~\cite{mobilealoha} & \redx & \redx & \redx & \greentick & \greentick & \redx & \greentick & \redx \\
BEHAVIOR-1K~\cite{behavior1k} & \redx & \redx & \greentick & \redx & \redx & \greentick & \greentick & \redx \\
ManiSkill-HAB~\cite{maniskillhab} & \redx & \redx & \redx & \greentick & \redx & \greentick & \redx & \redx \\
InfiniteWorld~\cite{infiniteworld} & \greentick & \redx & \greentick & \redx & \redx & \greentick & \redx & \redx \\
M$^2$Diffuser~\cite{m2diffuser} & \redx & \redx & \redx & \greentick & \redx & \greentick & \redx & \redx \\
RoboSuite~\cite{robosuite}  & \redx & \redx & \redx & \greentick & \greentick & \greentick & \redx & \greentick \\
BRMData~\cite{BRMData} & \redx & \redx & \redx & \greentick & \greentick & \redx & \greentick & \redx \\
RoboCasa~\cite{robocasa} & Some & \greentick & \redx & \greentick & Some & \greentick & \greentick & \greentick \\
DeMoBot~\cite{demobot} & \redx & \redx & \redx & \greentick & \greentick & \greentick & \greentick & \greentick \\
MoMaRT~\cite{momart}  & \redx & \redx & \redx & \greentick & \greentick & \greentick & \redx & \redx \\
M$^3$Bench~\cite{m3bench}  & \greentick & \redx & \redx & \greentick & \redx & \greentick & \redx & \redx \\
PARTNR~\cite{partnr} & \greentick & \greentick & \greentick & \redx & \redx & \greentick & \redx & \redx\\
BiGym~\cite{bigym} & \redx & \redx & \redx & \greentick & \greentick & \greentick & \redx & \redx \\
RT-1~\cite{rt1}  & \greentick & \redx & \redx & \greentick & \greentick & \greentick & \greentick & \redx \\
\hline
$\lambda$          & \greentick & \greentick & \greentick & \greentick & \greentick & \greentick & \greentick & \greentick \\
\hline
\end{tabular}
}
\caption{\textbf{Comparison of MoMa Benchmarks.}
$\lambda$ is the first MoMa benchmark that comprises human-collected demonstrations of language-conditioned long-horizon tasks in simulated and real-world environments utilizing a quadruped.
``NL" denotes Natural Language.
``FF" denotes Free-Form.
``Long-Horizon" is defined as long-horizon indoor navigation across multiple rooms and/or floors.
``Quadruped" refers to real quadrupeds.
``Some" indicates that only a portion of the benchmark possesses the respective quality or attribute.}
\label{tab:related_benchmarks}
\vspace{-7mm}
\end{table*}


\section{RELATED WORK}
As robot learning algorithms evolve rapidly, benchmarks have become critical for evaluating algorithmic decisions and ensuring consistent performance comparisons in standardized scenarios. 
However, existing benchmarks, shown in Table~\ref{tab:related_benchmarks}, fail to adequately evaluate models for data efficiency in long-horizon, language-conditioned indoor MoMa tasks, leaving significant gaps in advancing this area. 
For example, most benchmarks~\cite{humanoidbench, mobilealoha, behavior1k, maniskillhab, robosuite, BRMData, demobot, momart, bigym}, lack natural language annotations, despite its importance as an intuitive interface for specifying tasks, particularly in household and workplace scenarios. 
Language conditioning allows robots to interpret human commands directly, making this an essential feature for realistic applications.
Even fewer benchmarks, such as RoboCasa~\cite{robocasa} and PARTNR~\cite{partnr}, offer natural language instructions in a free-form, unconstrained format rather than predefined templates.
This flexibility is crucial, as it better reflects the variability and nuance of human communication that robots will encounter in the wild.

Moreover, almost all benchmarks~\cite{humanoidbench, mobilealoha, maniskillhab, robosuite, robocasa, BRMData, demobot, momart, m3bench, bigym, rt1, m2diffuser} do not support long-horizon tasks involving navigation across rooms and/or floors. 
These tasks are critical for indoor robotics, where robots must traverse long distances to perform daily operations. 
Without this capability, benchmarks fail to reflect the complexity of real-world scenarios. 
As shown in Table~\ref{tab:related_benchmarks}, Mobile ALOHA~\cite{mobilealoha} and BRMData~\cite{BRMData} lack simulation data entirely.
Simulated environments offer diverse scenes and objects, making it easier to test models across a wide range of scenarios.
Their standardized and controlled conditions create a portable testbed that others can readily use, ensuring consistency and reproducibility in evaluations.
On the other hand, most lack real-world data~\cite{safeagentbench, humanoidbench, maniskillhab, infiniteworld, robosuite, momart, m3bench, bigym, partnr, m2diffuser}, which is essential for providing a benchmark grounded in realistic scenarios where robots will ultimately be deployed, ensuring models are tested under practical and uncontrolled conditions.

The majority of benchmarks in Table~\ref{tab:related_benchmarks} provide expert demonstrations for the tasks, which are crucial for establishing ground-truth data to train learning-based methods and test models.
However, a few, SafeAgentBench~\cite{safeagentbench}, HumanoidBench~\cite{humanoidbench}, BEHAVIOR-1K~\cite{behavior1k}, InfiniteWorld~\cite{infiniteworld}, and PARTNR~\cite{partnr}, lack this attribute, limiting their utility for model training and evaluation.
M$^3$Bench~\cite{m3bench}, M$^2$Diffuser~\cite{m2diffuser}, and ManiSkill-HAB~\cite{maniskillhab} depend on planner-generated trajectories.
These lack the natural variability and expertise of human-collected demonstrations, which more accurately reflect real-world conditions and improve the quality of learned behaviors by providing nuanced, high-quality examples that enable better generalization.

Finally, as shown in Table~\ref{tab:related_benchmarks}, RoboCasa~\cite{robocasa}, RoboSuite~\cite{robosuite}, and DeMoBot~\cite{demobot} are the only benchmarks featuring armed real quadruped data, which is essential for tasks requiring stability and adaptability. 
Quadrupeds with manipulators can handle uneven terrains, such as stairs, while carrying objects, making them ideal for multi-floor environments, such as those found in $\lambda$.
Their robustness allows them to maintain stability in obstacle-dense environments, withstand collisions, and recover by picking themselves up if they fall, making them especially suitable for realistic and demanding scenarios.
The absence of this embodiment limits the applicability of the vast majority of benchmarks to realistic robotic use cases.

By addressing these limitations, $\lambda$ provides a comprehensive benchmark to test the data efficiency of MoMa models.
It integrates language-conditioned tasks, long-horizon trajectories spanning multiple rooms and floors, and both simulated and real-world data. 
Additionally, its human-collected demonstrations ensure realistic variability, and its inclusion of armed real quadruped data makes it uniquely suited for advancing robotics in complex settings.


\section{$\lambda$ BENCHMARK}
The LAMBDA ($\lambda$) benchmark is designed to evaluate model data efficiency on long-horizon MoMa tasks. 
$\lambda$ features language-conditioned, multi-room, multi-floor pick-and-place demonstrations spanning a range of diverse environments across simulated and real settings.

\subsection{Challenges of $\lambda$ Tasks}\label{subsec:A}
\textbf{Data Efficiency in MoMa.}
We define data efficiency as the ability to achieve relatively strong performance with minimal robot demonstrations, where each trajectory consists of a sequence of robot observations, states, and actions at every timestep.
Non-end-to-end systems are still considered within our definition of data efficiency.
For example, modular methods that integrate off-the-shelf Vision-Language Models (VLMs), which are not trained on robot demonstrations, are not excluded on this basis.
Unlike large MoMa datasets such as RT-1~\cite{rt1}, $\lambda$ is designed to test the data efficiency of learning-based models, done in Section~\ref{sec:eval}, by offering a dataset of manageable size, as shown in Table~\ref{tab:splits}.

While previous tabletop manipulation methods depended on thousands of demonstrations~\cite{vima}, recent models~\cite{gem, diffusionpolicy, transporter} are more data efficient, requiring only tens of demonstrations.
In a similar direction, recognizing the increased complexity of MoMa, we provide hundreds of demonstrations for model development and benchmarking.
This deliberate size balances diversity for learning with a manageable scale while also challenging models to be more data efficient in out-of-distribution generalization, avoiding the excessive resource costs of unrealistically large datasets.

\textbf{Long Horizon Tasks in Complex Spaces.}
MoMa tasks introduce significant complexity by involving larger and more intricate action, observation, state, and task spaces compared to tabletop manipulation.
Operating in environments that span multiple rooms and floors (Figure~\ref{fig:main_splash}), these tasks require the integration of high-level planning with low-level motion control.
For example, a robot may need to navigate to a dining table, pick up a plate, and transport it to a countertop in another room. 
Even in this seemingly straightforward pick-and-place scenario, the nature of the task manifests in various approaches—the robot might approach the table from different angles, select from multiple grasping points on the plate~\cite{handeyecoord}, or choose among several routes to the destination.  

\begin{figure}
    \centering
    \includegraphics[width=0.48\textwidth]{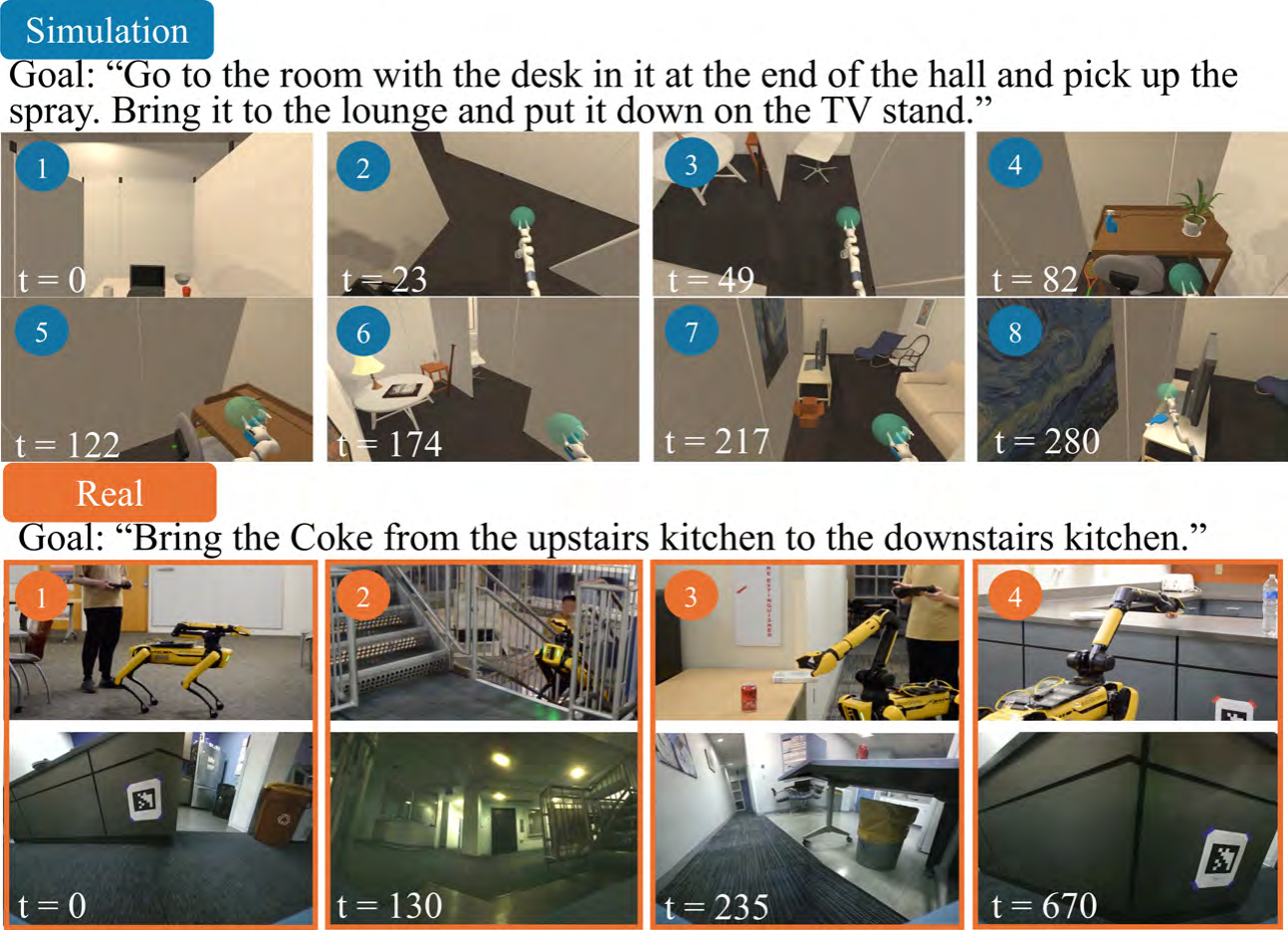}
    \caption{\textbf{Demonstrations.}
    Language-specified MoMa tasks with their expert demonstrations.
    The top orange row shows the teleoperation of Spot, and the bottom depicts its egocentric observation.}
    \label{fig:trajs}
    \vspace{-7mm}
\end{figure}

The challenges are further amplified in long-horizon tasks, where extended action sequences necessitate hierarchical strategies~\cite{feudal} that consider spatial relationships, temporal dependencies, and the compounding effects of errors over time—issues that are also observed in covariate shifts with BC approaches~\cite{stablebc}. 
Unlike fixed single-arm setups, MoMa's expansive spaces force the robot to contend with a higher degree of variability and uncertainty.
Consequently, our benchmark focuses on pick-and-place tasks (e.g., Figure~\ref{fig:trajs}), which serve as a representative foundation for addressing the broader demands of MoMa models in complex environments.

\textbf{Crowdsourced Free-Form Natural Language Instructions.}
Understanding language is crucial for robots in human environments, as it is the most natural way for humans to specify tasks.
Thus, $\lambda$'s tasks are language-conditioned, meaning goals are not specified via images, state information, or symbolic representations.
Language adds complexity, requiring models to interpret and ground linguistic expressions~\cite{nltellex} instead of relying on explicit visual or spatial cues.

To ensure tasks in $\lambda$ reflect real-world needs, we collected free-form human instructions via crowdsourcing.  
This provides a diverse and realistic set of tasks that people expect robots to perform in household and workplace settings.
Unlike system-generated templated commands, $\lambda$'s free-form instructions (e.g., Figures~\ref{fig:main_splash} \&~\ref{fig:trajs}) vary in phrasing, detail, and structure, requiring models to generalize across variations rather than memorizing fixed patterns. 
This pushes models to interpret and reason about task objectives.  
Additionally, these instructions introduce ambiguities and implicit references, further testing a model’s ability to align interpretations with human intent in long-horizon MoMa tasks.

\subsection{Simulated Dataset}
\label{subsec:B}
The dataset for our simulated tasks consists of $521$ trajectories, with an average length of $496$ timesteps, collected across $20$ rooms in five distinct environments, designed to ensure diversity in both object types and unique room layouts.
The object configurations within each environment are static.
Objects and furniture include, but are not limited to: clocks, bats, basketballs, apples, bowls, plates, forks, couches, TVs, chairs, desks, tables, trash cans, beds, shelves, drawers, lamps, etc.
We utilized the AI2THOR~\cite{ai2thor} simulator, specifically the RoboTHOR~\cite{robothor} environments, which feature connected rooms suitable for long-horizon tasks.

For robot control, we used ManipulaTHOR~\cite{manipulathor}, which includes an arm capable of fine manipulation and provides low-level pose data.
Each trajectory includes a range of attributes, including but not limited to RGB-D egocentric observations, segmentations, the poses of the robot body, end-effector, and grasped objects, as well as discrete actions.

To obtain realistic and diverse instructions, we used Prolific\footnote{\href{https://www.prolific.com}{\tt\small prolific.com}} for crowdsourcing to gather $521$ commands from $41$ participants, reflecting tasks that people would naturally expect robots to perform in indoor settings.
We teleoperated the robot with a keyboard-based teleoperation system we developed, to execute these commands.
Afterward, each trajectory was replay-verified to confirm correctness.

\subsection{Real-World Dataset}\label{subsec:C}
The dataset for our real-world tasks comprises $50$ trajectories collected across $11$ unique rooms in three distinct environments. 
These environments were chosen for their large size, thereby allowing long-horizon tasks, and their status as workplaces where robots can be particularly useful.
The first is a three-room laboratory, the second is a floor in a university building consisting of five rooms, and the third spans two adjacent floors connected by stairs within the same building, with eight rooms. 
These include kitchens, lobbies, a classroom, and a staircase, offering diverse objects and unique room layouts to challenge model generalizability.


To handle these complex terrains, we teleoperated the Spot~\cite{spot} robot using a tablet equipped with thumbsticks to collect task trajectories based on $50$ natural language commands provided by seven participants. 
The commands, distributed as $20$, $15$, and $15$ across the three environments, respectively, specified room-to-room and floor-to-floor tasks.
The trajectories were collected at $3$ Hz.


\begin{table*}[t]
\centering
\renewcommand{\arraystretch}{1} 
\resizebox{\textwidth}{!}{%
\begin{tabular}{l|ccc|ccc|ccc|ccc|ccc|ccc}
\toprule
 & \multicolumn{3}{c|}{\textbf{Scene 1}} & \multicolumn{3}{c|}{\textbf{Scene 2}} & \multicolumn{3}{c|}{\textbf{Scene 3}} & \multicolumn{3}{c|}{\textbf{Scene 4}} & \multicolumn{3}{c|}{\textbf{Scene 5}} & \multicolumn{3}{c}{\textbf{Task Gen}} \\ 
\cmidrule(r){2-16} \cmidrule(l){17-19}
\# Objects & \multicolumn{3}{c|}{17} & \multicolumn{3}{c|}{27} & \multicolumn{3}{c|}{16} & \multicolumn{3}{c|}{17} & \multicolumn{3}{c|}{18} & \multicolumn{3}{c}{-} \\
\# Rooms & \multicolumn{3}{c|}{4} & \multicolumn{3}{c|}{4} & \multicolumn{3}{c|}{3} & \multicolumn{3}{c|}{5} & \multicolumn{3}{c|}{4} & \multicolumn{3}{c}{-} \\
\hline
 & Train & Val & Test & Train & Val & Test & Train & Val & Test & Train & Val & Test & Train & Val & Test & Train & Val & Test \\ 
\midrule
Demos & 330 & 83 & 108 & 340 & 86 & 95 & 337 & 85 & 99 & 324 & 82 & 115 & 333 & 84 & 104 & 416 & 51 & 54 \\
Scenes & 4 & 4 & 1 & 4 & 4 & 1 & 4 & 4 & 1 & 4 & 4 & 1 & 4 & 4 & 1 & 5 & 5 & 5 \\
\bottomrule
\end{tabular}%
}
\caption{\textbf{Scene Information \& Data Splits.}
Splitting of the $521$ simulation demonstrations into subsets for the Scene and Task Generalization experiments.
The ``Scenes" row shows the number of scenes contributing to each subset of demonstrations.}
\label{tab:splits}
\vspace{-7mm}
\end{table*}


\section{EVALUATION}
\label{sec:eval}

\subsection{Baseline Models}
To evaluate the benchmark and establish its initial baselines, we test four MoMa systems.
Given the growing popularity of BC, we benchmark two state-of-the-art (SOTA) BC models. 
This provides insight into the data efficiency of current BC algorithms.

Additionally, we benchmark a neuro-symbolic system that integrates foundation models with task and motion planning (TAMP). 
The foundation models are pretrained on large-scale visual and textual data, but the system overall requires no robot demonstration training.
Thus, we evaluated it zero-shot to assess whether non-end-to-end methods can achieve greater data efficiency. 
By comparing its performance to learning-based approaches, this evaluation provides insight into whether such methods can achieve comparable or superior results, offering guidance for future development of data-efficient MoMa methodologies.
Finally, we assess the performance of a random agent to establish a lower-bound baseline, wherein actions are uniformly sampled.

\textbf{Robotics Transformer.}
RT-1~\cite{rt1} is a MoMa BC model.
It processes natural language instructions and RGB image inputs into discretized actions by leveraging a transformer~\cite{transformer}.
Its status as an open-source, SOTA learning-based MoMa model makes it suitable for benchmarking on $\lambda$ as a baseline.
To adapt RT-1 for the ManipulaTHOR robot, we modified its action head to align with the robot's capabilities. 
For task navigation, the robot base operates using discrete actions, while continuous deltas (discretized into 256 bins) are utilized for body rotation and manipulation. 

\textbf{MotionGlot-MoMa.} 
MotionGlot (MG)~\cite{motionglot} addresses the problem of varying action and observation dimensions in multi-embodiment models~\cite{openvla}. 
Leveraging training principles from language models, MG proposes a training procedure for motion generation across multiple embodiments as a next-token prediction problem.
We adapt MG for MoMa by leveraging $\lambda$'s demonstrations consisting of paired RGB observations and actions, calling it MG-MoMa.
MG-MoMa is trained to learn a language-conditioned model that maps those observations to corresponding actions.
This process involves two stages. 
First, a VQ-VAE~\cite{vqvae} learns a discrete latent codebook for the images, forming the observation vocabulary, while discrete binning is applied to tokenize the actions. 
Second, both the observation and action vocabularies are integrated with the text vocabulary from GPT-2~\cite{gpt2}. 
This unified vocabulary enables action generation akin to text generation, utilizing the instruction-tuning template from MG-MoMa for text command-based action execution.

\textbf{LIMP.}
Language Instruction grounding for Motion Planning (LIMP)~\cite{limp} is a neuro-symbolic system for pick-and-place MoMa integrating Linear Temporal Logic (LTL), a Large Language Model (LLM), VLM, and novel TAMP algorithms to process natural language instructions into executable TAMP plans. 
By translating instructions into LTL formulas enriched with Composable Referent Descriptors~\cite{limp}, LIMP dynamically generates context-aware specification translations and compiles them into finite-state automata for verifiably correct execution.
Unlike learning-based methods, LIMP requires no training on robot demonstrations, allowing it to be evaluated zero-shot. 
This enables a comparison of its data efficiency against BC models, providing insight into whether such systems are a promising direction for developing more data-efficient MoMa approaches.

\subsection{Experiments.}
\label{subsec:exp}
To evaluate data efficiency, we assess the models on scene generalization and task generalization, two widely used experiment types~\cite{rt1, demobot}.
These evaluations are essential for benchmarking data efficiency because they test whether models can learn effectively from limited data and apply that knowledge to novel scenarios. 
We define Scene Generalization as robots performing tasks in unseen environments with different layouts and/or objects, while Task Generalization as the ability to adapt to new tasks within seen settings. 
Together, these experiments quantify a model's data efficiency by isolating generalization challenges at both the scene and task levels, providing insights into how well models leverage limited training data to achieve robust performance.

\textbf{Data Split \& Generalization.}
To implement these evaluations, we conduct experiments using carefully designed data splits on the simulation dataset, highlighted in Table~\ref{tab:splits}.
For Scene Generalization, we employ a $5$-fold cross-validation approach, where each of the five scenes in the dataset is held out as the test scene in turn. 
The held-out test set consists of all trajectories from the designated scene, while the training and validation sets are derived from the remaining four scenes. 
Specifically, $80\%$ of the trajectories from each of the four remaining scenes are included in the training set, and $20\%$ are allocated to the validation set, following a stratified split to preserve data distribution.
During each fold, we train and hyperparameter-tune a separate model specific to the held-out test scene. 
This process ensures that we assess the model’s ability to generalize to entirely unseen environments while maintaining experimental controls.
The rotation of held-out scenes allows for a comprehensive evaluation of Scene Generalization across the entire dataset.

For Task Generalization, the data split is consistent across all five scenes. 
The training set comprises $80\%$ of the trajectories from each scene, the validation set includes $10\%$, and the test set includes the remaining $10\%$. 
Unlike Scene Generalization, the held-out test set in this experiment is fixed and evaluates the model’s ability to generalize to novel tasks rather than unseen environments. 
Hyperparameter tuning is performed, resulting in a single model optimized for Task Generalization on the test set.

\textbf{Training.}
We train RT-1 and MG-MoMa without pretrained parameters on the simulation tasks.
This evaluates how well they learn from our realistically sized dataset, providing a measure of their data efficiency specifically for long-horizon MoMa tasks.
Both models are trained using the Scene Generalization and Task Generalization experiments.

\textbf{Fine-Tuning.}
We fine-tune RT-1 and MG-MoMa with pretrained parameters on the simulation tasks.
Our empirical analysis showed that for RT-1, fine-tuning with some frozen layers led to better performance than not freezing any, so this approach was used for the final experiment. 
In contrast, for MG-MoMa, the unfrozen model performed better, leading to the use of fully trainable parameters.
This discrepancy may stem from MG-MoMa needing to override task-irrelevant biases from its original non-MoMa pretraining, necessitating updates across all layers.
In contrast, since RT-1 was already pretrained on MoMa tasks, it may retain useful representations relevant to $\lambda$ tasks in certain layers.
Using this setup, we conducted the Task Generalization experiment to measure the extent to which pretrained MoMa knowledge impacts the models' ability to adapt efficiently to new MoMa tasks.

\textbf{Metric.}
We use a task success rate metric that quantifies the completion of long-horizon MoMa tasks. 
Each task consists of four sequential dependent subtasks: (1) navigating to the target object, (2) grasping the object, (3) transporting the object to the goal location, and (4) placing the object at the goal. 
Each successful subtask contributes one point, with a maximum of four points per task. 
The success rate is calculated by averaging the subtask scores within each task and then averaging across all tasks.
This captures both navigation and manipulation performance, providing insight into a model’s ability to execute long-horizon tasks.



\begin{table}[t]
\centering
\renewcommand{\arraystretch}{1.65} 
\resizebox{\columnwidth}{!}{%
\begin{tabular}{l|cccccc}
\toprule
 & \multicolumn{6}{c}{\textbf{Scene Gen}} \\ \cmidrule(l){2-7}
\textbf{Model} & Scene 1 & Scene 2 & Scene 3 & Scene 4 & Scene 5 & Avg \\
\midrule
\textbf{RT-1} & $4.9\pm1.0 (10.0)$ & $0.78\pm0.5 (4.4)$ & $4.3\pm1.0 (9.5)$ & $2.0\pm0.6 (6.7)$ & $1.4\pm0.6 (5.9)$ & $2.7$ \\ 
\textbf{MG-MoMa} & $1.9\pm0.6 (5.8)$ & $3.7\pm0.9 (8.9)$ & $0.8\pm0.4 (4.3)$ & $2.0\pm0.6 (6.7)$ & $3.8\pm0.9 (9.1)$ & $2.44$ \\
\textbf{Random} & $0.7\pm0.4 (4.1)$ & $0.5\pm0.4 (3.6)$ & $0.3\pm0.3 (2.5)$ & $0.0\pm0.0 (0.0)$ & $0.0\pm0.0 (0.0)$ & $0.3$\\
\bottomrule
\end{tabular}%
}
\caption{\textbf{Scene Generalization.}
Success rates for Scene Generalization on models trained from scratch.
Parentheses indicate the standard deviation across all evaluated tasks.
}
\label{tab:scene_gen}
\vspace{-7mm}
\end{table}


\section{Discussion}

\subsection{Results} 

\textbf{Scene Generalization.}
The success rates for the Scene Generalization experiment, detailed in Table~\ref{tab:scene_gen}, are exceptionally low, with an overall average of just $2.7\%$ and $2.4\%$ across all scenes for RT-1 and MG-MoMa, respectively.
This demonstrates significant challenges in generalizing to novel scenes with unseen room layouts, furniture configurations, and object placements for our long-horizon MoMa tasks.
Both models outperform the random agent, demonstrating at least a minimal capacity to learn something from the data.

Scenes 1 and 3 exhibited relatively higher success rates compared to the others for RT-1. 
This can be attributed to a relatively high proportion of evaluation tasks in these scenes involving objects that were positioned directly in front of or adjacent to the robot's initial spawn location, reducing the complexity of navigation and manipulation required. 
In contrast, Scene 2 for RT-1 exhibited the lowest performance, likely due to the robot spawning in an enclosed and isolated room where many tasks necessitated navigating out of the room to retrieve the target object.
Additionally, Scene 2 contained the highest number of objects compared to the other scenes (see Figure~\ref{tab:splits}).
These factors combined significantly increased task complexity.
Overall, MG-MoMa demonstrates performance that is near RT-1, which is noteworthy given that MG was not originally intended for MoMa tasks.

\textbf{Task Generalization.}
Table~\ref{tab:task_gen} presents the results of our Task Generalization experiments.
Notably, both RT-1 and MG-MoMa show slightly higher overall performance compared to the Scene Generalization experiments. 
This improvement is likely due to the larger training dataset size (see Table~\ref{tab:splits}) and the availability of all scenes during training.
Moreover, while the performance of the baselines is equal for the seen settings, RT-1 outperforms MG-MoMa by $0.9\%$ on the unseen environments.
Meanwhile, the random agent showed no difference in performance between seen and unseen tasks, as expected.

\textbf{Task Generalization with Fine-Tuning.}
RT-1's unseen task performance was marginally lower than the fully trained model’s, as shown in Table~\ref{tab:task_gen}. 
Overall, seen and unseen results were similar, indicating that the models did not overfit.
However, MG-MoMa exhibited relatively lower performance on seen tasks compared to unseen tasks when fine-tuned.
This may mean that the pretrained parameters biased the models toward their original training tasks and away from effectively adapting to the unique challenges of $\lambda$, especially since the original models were trained on real-world data.



\begin{table}[t]
\centering
\renewcommand{\arraystretch}{1} 
\resizebox{\columnwidth}{!}{%
\begin{tabular}{l|cc|cc}
\toprule
 & \multicolumn{2}{c|}{\textbf{Task Gen}} & \multicolumn{2}{c}{\textbf{Task Gen FT}} \\ \cmidrule(r){2-3} \cmidrule(l){4-5}
\textbf{Model} & Seen & Unseen & Seen & Unseen \\
\midrule
\textbf{RT-1} & $4.6\pm1.3 (9.8)$ & $5.1\pm1.4 (10.2)$ & $4.6\pm1.3 (9.8)$ & $4.6\pm1.3 (9.8)$ \\ 
\textbf{MG-MoMa} & $4.6\pm1.3 (9.8)$ & $4.2\pm1.3 (9.4)$ & $3.7\pm1.2 (9.0)	$ & $4.5\pm1.3 (9.6)$ \\ 
\textbf{LIMP} & -- & $44.4\pm7.5 (50.3)^*$ & -- & -- \\
\textbf{Random} & $0.5\pm0.5 (3.4)$ & $0.5\pm0.5 (3.4)$ & -- & -- \\

\bottomrule
\end{tabular}%
}
\caption{\textbf{Task Generalization.}
Success rates for Task Generalization, including training from scratch and fine-tuning pretrained parameters.
``Seen" refers to evaluation on the training subset, while ``Unseen" applies to the test subset.
$^*$Indicates zero-shot evaluation without training, on a mix of simulation and real-world tasks.}
\label{tab:task_gen}
\vspace{-7mm}
\end{table}


\textbf{LIMP.} 
The results presented thus far show that learning-based models, BC in particular, face significant challenges in addressing the complexities of $\lambda$'s long-horizon MoMa tasks.
To explore an alternative paradigm, we benchmarked the LIMP system on the Task Generalization experiment.

The evaluation was conducted in real-world environments using Spot, as LIMP was originally designed for real-world execution.
We utilized $45$ tasks, drawn from a mix of simulated and real-world $\lambda$ tasks.
This mixed evaluation was necessary to test its performance on simulation-specified tasks as well.
This enabled us to evaluate LIMP’s versatility across both task types, providing insights into its adaptability to different environmental contexts.
For the simulation-designed tasks, real-world setups were constructed to approximate their simulated counterparts.  
As shown in Table~\ref{tab:task_gen}, LIMP achieved a relatively impressive average success rate of $44.4\%$ on unseen tasks, vastly outperforming RT-1 and MG-MoMa without any training on robot demonstrations, making it relatively more data efficient.

One key factor contributing to LIMP’s success is its ability to generate instruction-conditioned semantic maps of the environment.
These maps provide the robot with precise information about the locations of objects and regions of interest, eliminating the need to search and enabling the interpretation of fine-grained semantics specified in language-conditioned tasks.
Additionally, the maps allow the robot to utilize robust motion planning algorithms for navigation, which are less susceptible to covariate shift compared to the mapless action decoding used in BC models like RT-1.  
This robustness is critical for the tasks' long-horizon navigation aspect, where the inherent large observation and state spaces exacerbate the challenges faced by learning-based methods.
Another significant advantage of LIMP is its incorporation of multimodal foundation models, which endow it with world knowledge to reason about task requirements and perceptual inputs in a zero-shot manner. 
This capability enables it to handle the semantic and contextual complexities of our benchmark tasks without requiring task-specific training.

While LIMP demonstrates considerable promise, it still falls short of perfect execution, indicating room for improvement in handling the intricacies of our long-horizon MoMa tasks.
In particular, $\lambda$ revealed that LIMP exhibits an object-centric bias; it lacks an explicit understanding of spatial constructs such as rooms (e.g., it can navigate to a ``cup" but not to a ``kitchen"). 
It also lacks a notion of the starting location, rendering it unable to interpret instructions that require returning to its point of origin. 
Moreover, due to its reliance on a VLM, its performance is constrained by the VLM's perceptual limitations, which in this case led to failures in recognizing or locating target objects.
Overall, LIMP represents a promising direction for future MoMa development, blending the strengths of learning-based approaches with those of classical planning.


\begin{figure}
    \centering
    \includegraphics[width=0.48\textwidth]{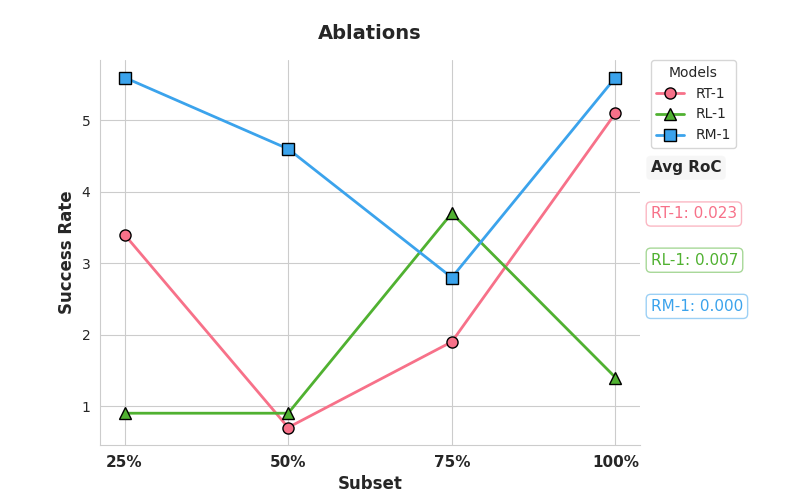}
    \caption{\textbf{Ablations.}
    Success rates for Task Generalization on growing subsets of the simulation dataset.
    ``Avg RoC": average rate of change; ``RL-1": RT-1 with LSTM; ``RM-1": RT-1 with Mamba.
    }
    \label{fig:ablation_graph}
    \vspace{-7mm}
\end{figure}


\subsection{Ablations}
\textbf{Impact of Dataset Size.}
ALOHA Unleashed~\cite{alohaunleashed}, a bimanual manipulation system trained on $26$K demonstrations, only achieved a success rate of $25\%$ for certain tasks, showing that even with large datasets, high performance is not guaranteed.
Similarly, Lin et al.~\cite{datascalinglaws} found that increasing demonstrations generally improves manipulation performance, though with diminishing returns.
Since our benchmark involves MoMa—with larger action, state, and observation spaces due to added navigation—and uses far fewer trajectories, substantial performance improvements likely require more data.
To investigate this, we pose the following questions:
Are the baseline models data inefficient?
Will scaling the data improve their performance?
To answer these, we investigate how performance scales when progressively larger subsets of the simulation dataset ($25\%$, $50\%$, $75\%$, and $100\%$) are used for training in the Task Generalization experiment.
This evaluates whether these models can effectively utilize limited data or require more, testing their data efficiency and scalability for long-horizon MoMa tasks.

The ablation results, summarized in Figure~\ref{fig:ablation_graph}, show a positive upward trend in performance for RT-1 (represented by the pink line) as the training data increases, albeit a weak one. 
This trend is evident in its average rate of change (RoC).
This shows that scaling the data does improve RT-1's overall performance.
This pattern is consistent with what is observed in the language, vision, and multimodal communities~\cite{scalingbillions,  revisitingscaling, mixedmodelsscaling}.
Our results reflect that these BC models are data inefficient for long-horizon MoMa tasks, as the less data, the worse they generally perform.
This raises questions about the source of inefficiency, such as model architecture.

\textbf{Impact of Architecture.}
To assess whether data inefficiency stems from the architecture, we evaluate and compare different architectures.
While MG-MoMa was already benchmarked and provides some architectural comparison with RT-1, it is not sufficiently different since it also uses a transformer~\cite{transformer}.
To explore distinct alternatives, we replace RT-1's transformer with alternate architectures while keeping all other components, such as encoders, unchanged to ensure a controlled comparison.
Specifically, we replace the transformer with an LSTM~\cite{lstm}, and refer to the model as RL-1. 
Likewise, we incorporate the relatively new Mamba architecture~\cite{mamba}, and call this RM-1.
To understand how these alternative architectures perform with increasing data, we evaluate both RL-1 and RM-1 under the Task Generalization setup using the same subset-based evaluation previously conducted.
By comparing their trends/RoCs, subsets, and full dataset results against RT-1's, we aim to determine whether data inefficiency is tied to architectural limitations.

As presented in Figure~\ref{fig:ablation_graph}, both RT-1 and RL-1 exhibit upward trends with similar RoCs.
This indicates that RL-1 did not learn more efficiently than RT-1.
However, RM-1 maintained a relatively constant trend and outperformed RT-1 across all subsets and the full dataset.
This finding shows that architecture could be a contributing factor in the observed data inefficiency associated with $\lambda$-type tasks.
Identifying more concrete and exact causes of data inefficiency requires further investigation, which is beyond the scope of this work and is left for the community to explore.

\section{Conclusion}


We introduce LAMBDA ($\lambda$), a challenging benchmark consisting of a realistic amount, $571$, of language-conditioned, long-horizon MoMa tasks, spanning room-to-room and floor-to-floor navigation in indoor environments. 
Each task is paired with human-collected expert demonstrations from both simulated and real-world settings.

To evaluate the benchmark, we tested baseline models to assess their performance and data efficiency on these complex tasks. 
End-to-end learning-based models, RT-1 and MG-MoMa, demonstrated poor performance and were found to be data-inefficient, highlighting limitations in their ability to generalize to $\lambda$'s long-horizon tasks. 
In contrast, a neuro-symbolic system, LIMP, significantly outperformed the BC models, showcasing its potential as a promising direction for the development of future MoMa systems capable of addressing the challenges posed by $\lambda$'s tasks.
We envision this benchmark as a valuable resource for the robotics community, highlighting data efficiency limitations in models and guiding the development of more efficient MoMa systems.

\textbf{Limitations and Future Work.} 
Incorporating a quadruped in real-world tasks showcases the advantages of quadrupedal platforms, but the absence of simulated quadruped data limits exposure to diverse environments, such as varied floor-to-floor trajectories.
Expanding the benchmark to include that would enhance generalization and enrich evaluation datasets.
In the future, $\lambda$ could be extended to include tasks beyond pick-and-place, such as those involving more intricate manipulation.

\section*{ACKNOWLEDGMENT}
This work is supported by ONR under grant award numbers N00014-22-1-2592 and N00014-23-1-2794, NSF under grant award number CNS-2150184, and with support from Amazon Robotics.
We thank Aryan Singh, George Chemmala, Ziyi Yang, David Paulius, Ivy He, Lakshita Dodeja, Mingxi Jia, Benned Hedegaard, Thao Nguyen, Tuluhan Akbulut, Chaerin Min, Selena Williams, and George Konidaris.





\footnotesize{
\bibliographystyle{IEEEtran} 
\bibliography{IEEEabrv}
}

\end{document}